\documentclass[10pt,twocolumn,letterpaper]{article}
\pdfoutput=1
\usepackage{iccv}
\usepackage{times}
\usepackage{epsfig}
\usepackage{graphicx}
\usepackage{amsmath}
\usepackage{amssymb}

\usepackage{subfigure}
\makeatletter
\g@addto@macro\normalsize{%
  \abovedisplayskip 2pt plus1pt minus2pt%
  \belowdisplayskip \abovedisplayskip
  \abovedisplayshortskip  0pt plus2pt%
  \belowdisplayshortskip  2pt plus1pt minus2pt%
\abovecaptionskip  1pt %
\belowcaptionskip  1pt %
}

\usepackage{cite}
\usepackage[pagebackref=true,breaklinks=true,letterpaper=true,colorlinks,bookmarks=false]{hyperref}

\iccvfinalcopy 

\def\iccvPaperID{1029} 

\begin{document}

\title{Beyond Face Rotation: Global and Local Perception GAN for Photorealistic and Identity Preserving Frontal View Synthesis}
\author{
Rui Huang$^{1,2}$\thanks{These two authors contributed equally.}\hspace{4pt}$\thanks{ Homepage http://andrew.cmu.edu/user/ruih2/} \hspace{10pt}$ Shu Zhang$^{1,2,3}$\footnotemark[1] \hspace{10pt}Tianyu Li$^{1,2}$ \hspace{10pt}Ran He$^{1,2,3}$\\
$^{1}$National Laboratory of Pattern Recognition, CASIA\\
$^{2}$Center for Research on Intelligent Perception and Computing, CASIA\\
$^{3}$University of Chinese Academy of Sciences, Beijing, China\\
{\tt\small huangrui@cmu.edu, tianyu.lizard@gmail.com, \{shu.zhang, rhe\}@nlpr.ia.ac.cn}
}

\maketitle

\begin{abstract}

Photorealistic frontal view synthesis from a single face image has a wide range of applications in the field of face recognition. Although data-driven deep learning methods have been proposed to address this problem by seeking solutions from ample face data, this problem is still challenging because it is intrinsically ill-posed. This paper proposes a Two-Pathway Generative Adversarial Network (TP-GAN) for photorealistic frontal view synthesis by simultaneously perceiving global structures and local details. Four landmark located patch networks are proposed to attend to local textures in addition to the commonly used global encoder-decoder network. Except for the novel architecture, we make this ill-posed problem well constrained by introducing a combination of adversarial loss, symmetry loss and identity preserving loss. The combined loss function leverages both frontal face distribution and pre-trained discriminative deep face models to guide an identity preserving inference of frontal views from profiles. Different from previous deep learning methods that mainly rely on intermediate features for recognition, our method directly leverages the synthesized identity preserving image for downstream tasks like face recognition and attribution estimation. Experimental results demonstrate that our method not only presents compelling perceptual results but also outperforms state-of-the-art results on large pose face recognition.

\end{abstract}

\section{Introduction}\label{sec:intro}

Benefiting from the rapid development of deep learning methods and the easy access to a large amount of annotated face images, unconstrained face recognition techniques~\cite{taigman2014deepface,sun2014deep} have made significant advances in recent years. Although surpassing human performance has been achieved on several benchmark datasets~\cite{schroff2015facenet}, pose variations are still the bottleneck for many real-world application scenarios. Existing methods that address pose variations can be divided into two categories. One category tries to adopt hand-crafted or learned pose-invariant features~\cite{schroff2015facenet,chen2013blessing}, while the other resorts to synthesis techniques to recover a frontal view image from a large pose face image and then use the recovered face images for face recognition~\cite{zhu2013deep,zhu2014multi}.

\begin{figure}[t]
\centering
\includegraphics[width=0.85\linewidth]{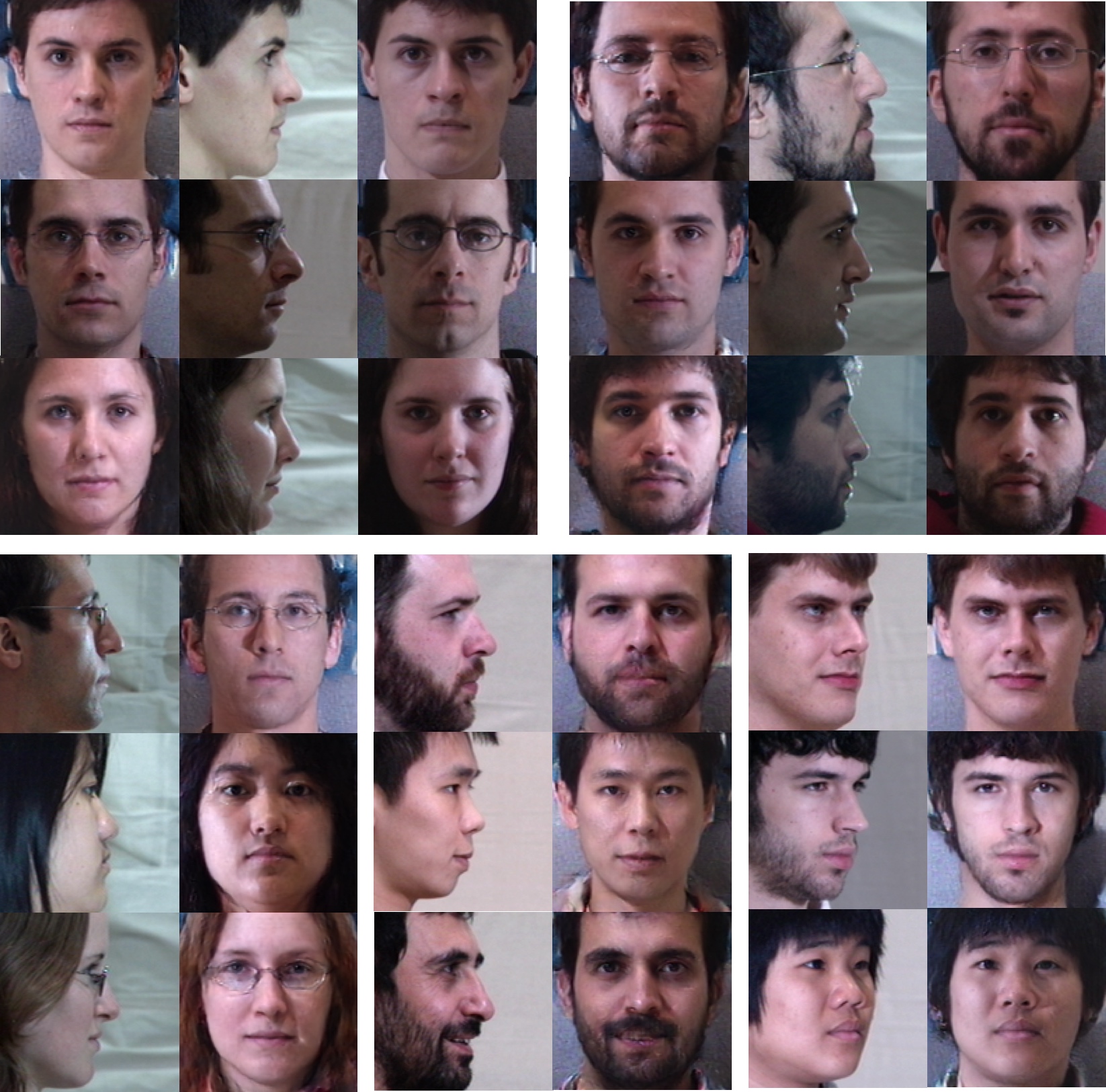} 
   \caption{Frontal view synthesis by TP-GAN. The upper half shows the $90^\circ$ profile image (middle) and its corresponding synthesized and ground truth frontal face. We invite the readers to guess which side is our synthesis results (please refer to Sec.~\ref{sec:intro} for the answer). The lower half shows the synthesized frontal view faces from profiles of $90^{\circ}$, $75^{\circ}$ and $45^{\circ}$ respectively.}
\label{fig:fongtpage}
\end{figure}

For the first category, traditional methods often make use of robust local descriptors such as Gabor~\cite{daugman1985uncertainty}, Haar~\cite{viola2001rapid} and LBP~\cite{ahonen2006face} to account for local distortions and then adopt metric learning~\cite{weinberger2009distance,chen2013blessing} techniques to achieve pose invariance.
In contrast, deep learning methods often handle position variances with pooling operation and employ triplet loss~\cite{schroff2015facenet} or contrastive loss~\cite{sun2014deep} to ensure invariance to very large intra-class variations. However, due to the tradeoff between invariance and discriminability, these approaches cannot deal with large pose cases effectively.

For the second category, earlier efforts on frontal view synthesis usually utilize 3D geometrical transformations to render a frontal view by first aligning the 2D image with either a general~\cite{hassner2015effective} or an identity specific~\cite{taigman2014deepface,zhu2015high} 3D model. 
These methods are good at normalizing small pose faces, but their performance decreases under large face poses due to severe texture loss. Recently, deep learning based methods are proposed to recover a frontal face in a data-driven way. For instance, Zhu \etal~\cite{zhu2014multi} propose to disentangle identity and pose representations while learning to estimate a frontal view. Although their results are encouraging, the synthesized image sometimes lacks fine details and tends to be blurry under a large pose so that they only use the intermediate features for face recognition. The synthesized image is still not good enough to perform other facial analysis tasks, such as forensics and attribute estimation.






Moreover, from an optimization point of view, recovering the frontal view from incompletely observed profile is an ill-posed or under-defined problem, and there exist multiple solutions to this problem if no prior knowledge or constraints are considered. Therefore, the quality of recovered results heavily relies on the prior or the constraints exploited in the training process. Previous work~\cite{zhu2013deep,zhu2014multi,yim2015rotating,kan2014stacked} usually adopts pairwise supervision and seldom introduce constraints in the training process, so that they tend to produce blurry results. 


When human try to conduct a view synthesis process, we firstly infer the global structure (or a sketch) of a frontal face based on both our prior knowledge and the observed profile. Then our attention moves to the local areas where all facial details will be filled out.
Inspired by this process, we propose a deep architecture with two pathways (TP-GAN) for frontal view synthesis. These two pathways focus on the inference of global structure and the transformation of local texture respectively. Their corresponding feature maps are then fused for further process for the generation of the final synthesis. We also make the recovery process well constrained by incorporating prior knowledge of the frontal faces' distribution with a Generative Adversarial Network (GAN)~\cite{goodfellow2014generative}.
The outstanding capacity of GAN in modeling 2D data distribution has significantly advanced many ill-posed low level vision problems, such as super-resolution~\cite{ledig2016photo} and inpainting~\cite{pathak2016context}. 
Particularly, 
drawing inspiration from the faces' symmetric structure, a symmetry loss is proposed to fill out occluded parts. Moreover, to faithfully preserve the most prominent facial structure of an individual, we adopt a perceptual loss~\cite{johnson2016perceptual} in the compact feature space in addition to the pixel-wise L1 loss. Incorporating the identity preserving loss is critical for a faithful synthesis and greatly improves its potential to be applied to face analysis tasks. We show some samples generated by TP-GAN in the upper half of Fig.~\ref{fig:fongtpage} (the left side of each tuple).

The main contributions of our work lie in three folds: 1) We propose a human-like global and local aware GAN architecture for frontal view synthesis from a single image, which can synthesize photorealistic and identity preserving frontal view images even under a very large pose. 2) We combine prior knowledge from data distribution (adversarial training) and domain knowledge of faces (symmetry and identity preserving loss) to exactly recover the lost information inherent in projecting a 3D object into a 2D image space. 3) We demonstrate the possibility of a ``recognition via generation'' framework and outperform state-of-the-art recognition results 
under a large pose. Although some deep learning methods have been proposed for face synthesis, our method is the first attempt to be effective for the recognition task with synthesized faces.

\begin{figure*}[t]
\centering

\includegraphics[width=0.7\linewidth]{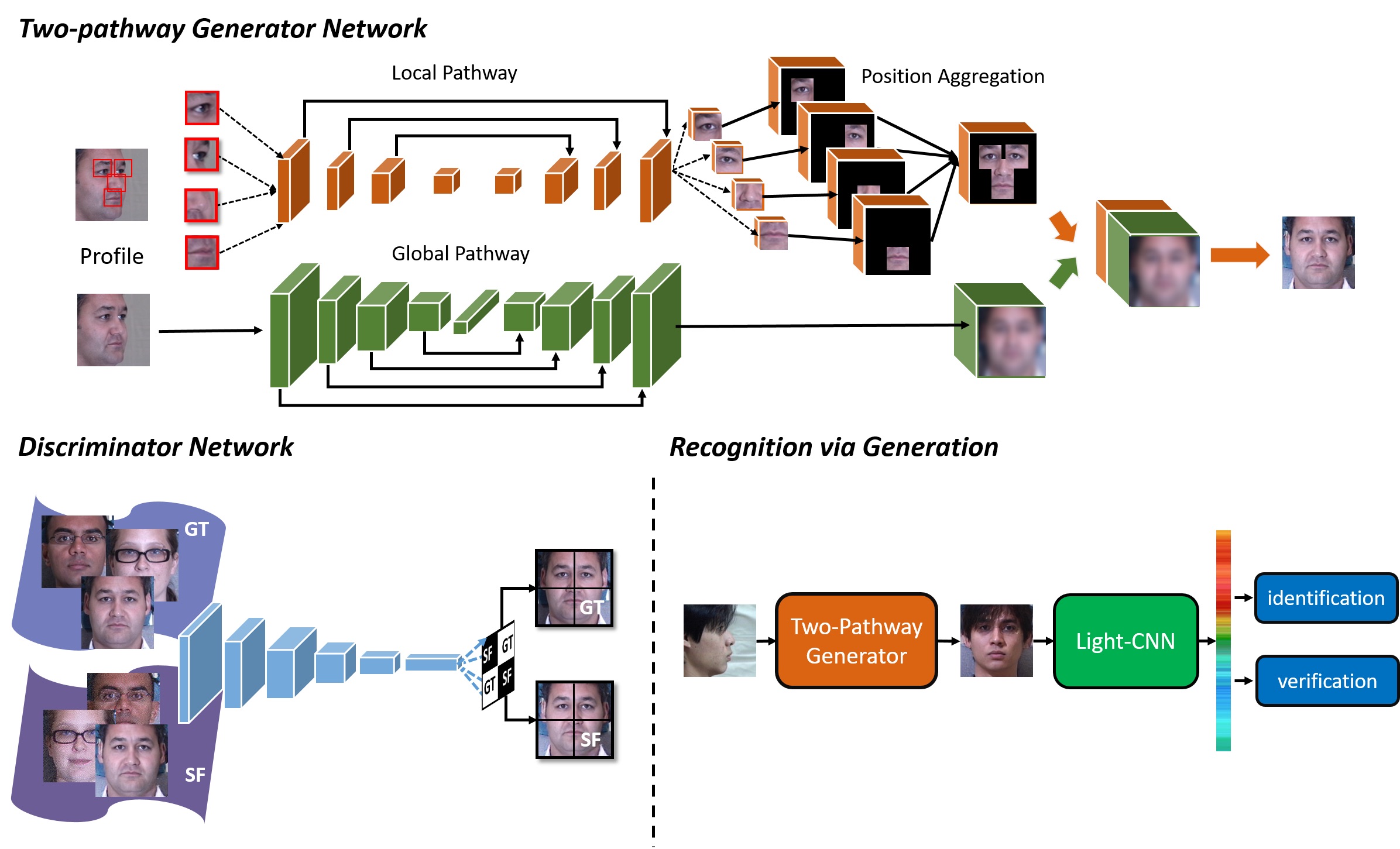}

   \caption{General framework of TP-GAN. The Generator contains two pathways with each processing global or local transformations. The Discriminator distinguishes between synthesized frontal (SF) views and ground-truth (GT) frontal views. Detailed network architectures can be found in the supplementary material.}
\label{fig:framework}
\end{figure*}

\section{Related Work}


\subsection{Frontal View Synthesis} Frontal view synthesis, or termed as face normalization, is a challenging task due to its ill-posed nature. Traditional methods address this problem either with 2D/3D local texture warping~\cite{hassner2015effective,zhu2015high} or statistical modeling~\cite{sagonas2015robust}. For instance, 
Hassner \etal~\cite{hassner2015effective} employ a mean 3D model for face normalization.
A joint frontal view synthesis and landmark localization method is proposed in~\cite{sagonas2015robust} with a constrained low-rank minimization model. Recently, researchers employ Convolutional Neural Networks (CNN) for joint representation learning and view synthesis~\cite{zhu2013deep,zhu2014multi,yim2015rotating,kan2014stacked}. Specifically, Yim \etal~\cite{yim2015rotating} propose a multi-task CNN to predict identity preserving rotated images. Zhu \etal~\cite{zhu2013deep,zhu2014multi} develop novel architectures and learning objectives to disentangle the identity and pose representation while estimating the frontal view. Reed \etal~\cite{honglak} propose to use a Boltzmann machine to model factors of variation and generate rotated images via pose manifold traversal. Although it is much more convenient if the synthesized image can be directly used for facial analysis tasks, most of the previous methods mainly employ intermediate features for face recognition because they cannot faithfully produce an identity preserving synthesis.

\subsection{Generative Adversarial Network (GAN)} As one of the most significant improvements on the research of deep generative models~\cite{kingma2013auto,rezende2014stochastic}, GAN~\cite{goodfellow2014generative} has drawn substantial attention from both the deep learning and computer vision society. The min-max two-player game 
provides a simple yet powerful way to estimate target distribution and generate novel image samples~\cite{denton2015deep}. With its power for distribution modeling, the GAN can encourage the generated images to move towards the true image manifold and thus generates photorealistic images with plausible high frequency details. Recently, modified GAN architectures, conditional GAN~\cite{mirza2014conditional} in particular, have been successfully applied to vision tasks like image inpainting~\cite{pathak2016context}, super-resolution~\cite{ledig2016photo}, style transfer~\cite{li2016combining}, face attribute manipulation~\cite{shen2016learning} and even data augmentation for boosting classification models~\cite{shrivastava2016learning,zheng2017unlabeled}. These successful applications of GAN motivate us to develop frontal view synthesis methods based on GAN.  

\section{Approach}

The aim of frontal view synthesis is to recover a photorealistic and identity preserving frontal view image $I^F$ from a face image under a different pose, \ie a profile image $I^P$. To train such a network, pairs of corresponding $\{I^F, I^P\}$ from multiple identities $y$ are required during the training phase. Both the input $I^P$ and output $I^F$ come from a pixel space of size $W \times H \times C$ with $C$ color channel.

It's our goal to learn a synthesis function that can infer the corresponding frontal view from any given profile images. Specifically, we model the synthesis function with a two-pathway CNN $G_{\theta_G}$ that is parametrized by $\theta_G$. Each pathway contains an Encoder and a Decoder, denoted as $\{G_{\theta ^g_E},  G_{\theta ^g_D}\}$ and $\{G_{\theta ^l_E},  G_{\theta ^l_D}\}$, where $g$ and $l$ stand for the global structure pathway and the local texture pathway respectively. In the global pathway, the bottleneck layer, which is the output of $G_{\theta ^g_E}$, is usually used for classification task~\cite{yang2015weakly} with the cross-entropy loss $L_{cross\_entropy}$.

The network's parameters $G_{\theta_G}$ are optimized by minimizing a specifically designed synthesis loss $L_{syn}$ and the aforementioned $L_{cross\_entropy}$. For a training set with $N$ training pairs of $\{I^F_n, I^P_n\}$, the optimization problem can be formulated as follows:

\begin{equation}
\begin{split}
\hat \theta_G = \frac{1}{N}\underset{\theta_G}{\mathrm{argmin}} \sum_{n=1}^{N} \{L_{syn}(G_{\theta_G}(I^P_n), I^F_n)\\
 + \alpha L_{cross\_entropy}(G_{\theta ^g_E}(I^P_n),y_n)\}
\label{eq:jointLoss}
\end{split}
\end{equation}
where $\alpha$ is a weighting parameter and $L_{syn}$ is defined as a weighted sum of several losses that jointly constrain an image to reside in the desired manifold. We will postpone the detailed description of all the individual loss functions to Sec.~\ref{subsec:synthesisfunction}.

\subsection{Network Architecture}

\subsubsection{Two Pathway Generator}

The general architecture of TP-GAN is shown in Fig.~\ref{fig:framework}. Different from previous methods~\cite{zhu2013deep,zhu2014multi,yim2015rotating,kan2014stacked} that usually model the synthesis function with one single network, our proposed generator $G_{\theta_G}$ has two pathways, with one global network $G_{\theta^g}$ processing the global structure and four landmark located patch networks $G_{\theta^l_i}, i\in\{0,1,2,3\}$ attending to local textures around four facial landmarks.

We are not the first to employ the two pathway modeling strategy. Actually, this is a quite popular routine for 2D/3D local texture warping~\cite{hassner2015effective,zhu2015high} methods. Similar to the human cognition process, they usually divide the normalization of faces into two steps, with the first step to align the face globally with a 2D or 3D model and the second step to warp or render local texture to the global structure. Moreover, Mohammed \etal~\cite{mohammed2009visio} combines a global parametric model with a local non-parametric model for novel face synthesis.

Synthesizing a frontal face $I^{F}$ from a profile image $I^{P}$ is a highly non-linear transformation. Since the filters are shared across all the spatial locations of the face image, we argue that using merely a global network cannot learn filters that are suitable for both rotating a face and precisely recovering local details. 
Therefore, we transfer the success of the two pathway structure in traditional methods to a deep learning based framework and introduce the human-like two pathway generator for frontal view synthesis.

As shown in Fig.~\ref{fig:framework}, $G_{\theta^g}$ is composed of a down-sampling Encoder $G_{\theta ^g_E}$ and an up-sampling Decoder $G_{\theta ^g_D}$, extra skip layers are introduced for multi-scale feature fusion. The bottleneck layer in the middle outputs a 256-dimension feature vector $v_{id}$, which is used for identity classification to allow for identity-preserving synthesis. At this bottleneck layer, as in~\cite{tran17}, we concatenate a 100-dim Gaussian random noise to $v_{id}$ to model variations other than pose and identity.

\subsubsection{Landmark Located Patch Network}


The four input patches of the landmark located patch network $G_{\theta ^l}$ are center-cropped from four facial landmarks, \ie left eye center, right eye center, nose tip and mouth center. Each $G_{\theta ^l_i}, i \in \{0,1,2,3\}$ learns a separate set of filters for rotating the center-cropped patch to its corresponding frontal view (after rotation, the facial landmarks are still in the center). The architecture of the landmark located patch network is also based on an encoder-decoder structure, but it has no fully connected bottleneck layer.

To effectively integrate the information from the global and local pathways, we adopt an intuitive method for feature map fusion. As shown in Fig.~\ref{fig:framework}, we firstly fuse the output feature tensors (multiple feature maps) of four local pathways to one single feature tensor that is of the same spatial resolution as the global feature tensor. Specifically, we put each feature tensor at a ``template landmark location", and then a max-out fusing strategy is introduced to reduce the stitching artifacts on the overlapping areas. Then, we simply concatenate the feature tensor from each pathway to produce a fused feature tensor and then feed it to successive convolution layers to generate the final synthesis output.

\subsubsection{Adversarial Networks}

To incorporate prior knowledge of the frontal faces' distribution into the training process, we further introduce an discriminator $D_{\theta_D}$ to distinguish real frontal face images $I^{F}$ from synthesized frontal face images $G_{\theta_G}(I^{P})$, following the work of Goodfellow \etal~\cite{goodfellow2014generative}. We train $D_{\theta_D}$ and $G_{\theta_G}$ in an alternating way to optimize the following min-max problem:
\begin{equation}
\begin{split}
\underset{\theta_G}\min \hspace{1pt} \underset{\theta_D}\max \hspace{1pt}
\mathbb{E}_{I^{F} \sim P(I^{F})} \log D_{\theta_D}(I^{F}) + \\
\mathbb{E}_{I^{P} \sim P(I^{P})}\log (1-D_{\theta_D}(G_{\theta_G}(I^{P})))
\end{split}
\end{equation}

Solving this min-max problem will continually push the output of the generator to match the target distribution of the training frontal faces, thus it encourages the synthesized image to reside in the manifold of frontal faces, leading to photorealistic synthesis with appealing high frequency details. As in~\cite{shrivastava2016learning}, our $D_{\theta_D}$ outputs a $2\times 2$ probability map instead of one scalar value. Each probability value now corresponds to a certain region instead of the whole face, and $D_{\theta_D}$ can specifically focus on each semantic region.

\subsection{Synthesis Loss Function}\label{subsec:synthesisfunction}
The synthesis loss function used in our work is a weighted sum of four individual loss functions, we will give a detailed description in the following sections.

\subsubsection{Pixel-wise Loss}

We adopt pixel-wise L1 loss at multiple locations to facilitate multi-scale image content consistency:

 \begin{equation}
\begin{split}
L_{pixel}= \frac{1}{W\times H}\sum_{x=1}^{W}\sum_{y=1}^{H}
\mathopen|I^{pred}_{x,y}-I^{gt}_{x,y} \mathclose|\\
\end{split}
\end{equation}

Specifically, the pixel wise loss is measured at the output of the global, the landmark located patch network and their final fused output. To facilitate a deep supervision, we also add the constraint on multi-scale outputs of the $G_{\theta ^g_D}$. Although this loss will lead to overly smooth synthesis results, it is still an essential part for both accelerated optimization and superior performance.

\subsubsection{Symmetry Loss}

Symmetry is an inherent feature of human faces. Exploiting this domain knowledge as a prior and imposing a symmetric constraint on the synthesized images may effectively alleviate the self-occlusion problem and thus greatly improve performance for large pose cases. Specifically, we define a symmetry loss in two spaces, \ie the original pixel space and the Laplacian image space, which is robust to illumination changes. 
The symmetry loss of a face image takes the form:

\begin{figure*}[t]
\centering

\subfigure[Profile]{
\begin{minipage}[b]{0.07\linewidth}
    \centering
   \includegraphics[width=1\linewidth]{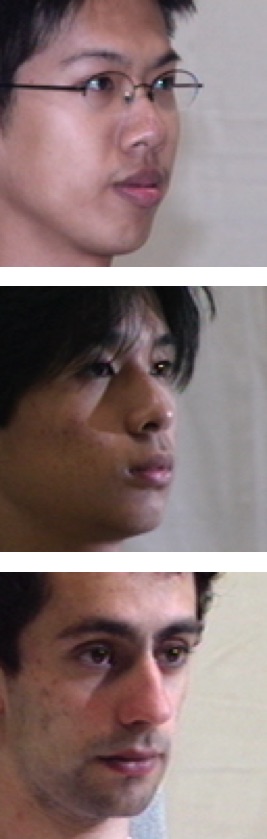}\vspace{0.03\linewidth}
\end{minipage}}
\subfigure[Ours]{
\begin{minipage}[b]{0.07\linewidth}
    \centering
   \includegraphics[width=1\linewidth]{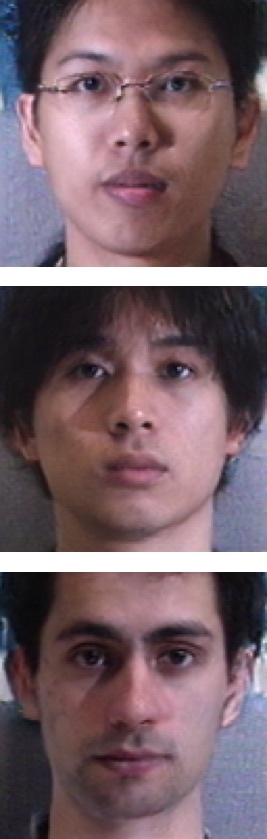}\vspace{0.03\linewidth}
\end{minipage}}
\subfigure[~\cite{tran17}]{
\begin{minipage}[b]{0.0715\linewidth}
    \centering
   \includegraphics[width=1\linewidth]{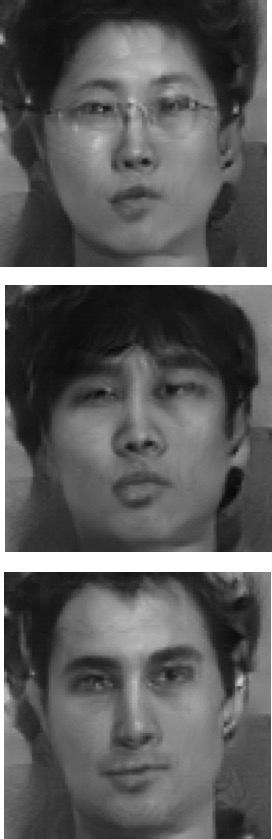}\vspace{0.03\linewidth}
\end{minipage}}
\subfigure[~\cite{yim2015rotating}]{
\begin{minipage}[b]{0.071\linewidth}
    \centering
   \includegraphics[width=1\linewidth]{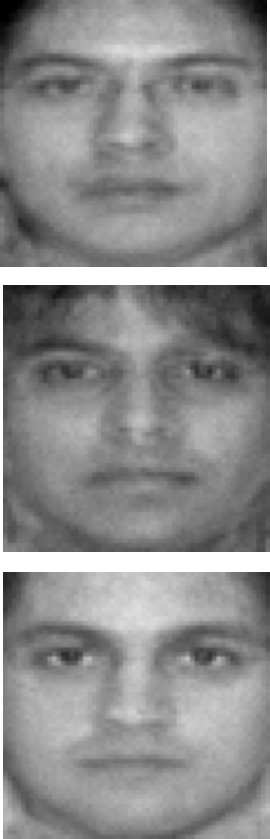}\vspace{0.03\linewidth}
\end{minipage}}
\subfigure[~\cite{face_bmvc16}]{
\begin{minipage}[b]{0.071\linewidth}
    \centering
   \includegraphics[width=1\linewidth]{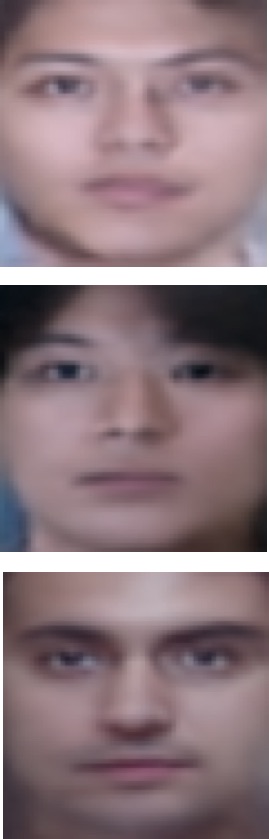}\vspace{0.03\linewidth}
\end{minipage}}
\subfigure[~\cite{zhu2015high}]{
\begin{minipage}[b]{0.07\linewidth}
    \centering
   \includegraphics[width=1\linewidth]{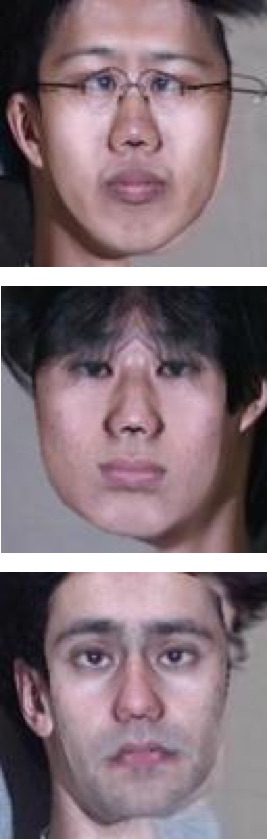}\vspace{0.03\linewidth}
\end{minipage}}
\subfigure[~\cite{hassner2015effective}]{
\begin{minipage}[b]{0.07\linewidth}
    \centering
   \includegraphics[width=1\linewidth]{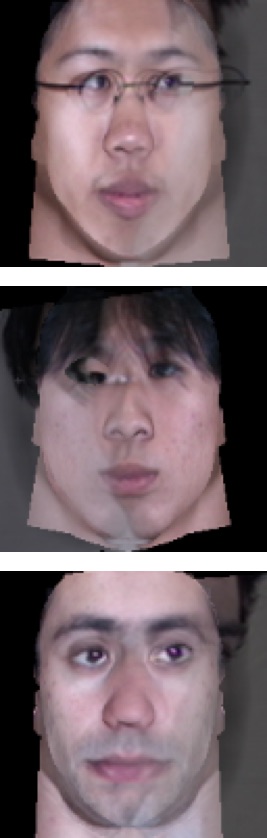}\vspace{0.03\linewidth}
\end{minipage}}
\subfigure[Frontal]{
\begin{minipage}[b]{0.07\linewidth}
    \centering
   \includegraphics[width=1\linewidth]{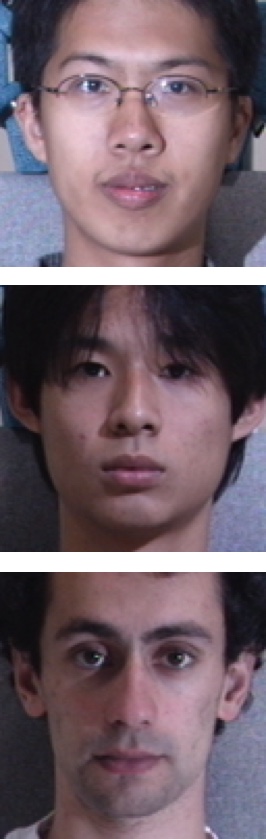}\vspace{0.03\linewidth}
\end{minipage}}
\caption{Comparison with state-of-the-art synthesis methods under the pose of $45^\circ$ (first two rows) and $30^\circ$ (last row). }
\label{fig:othermethods} 
\end{figure*}

\begin{equation}
\begin{split}
L_{sym} = \frac{1}{W/2\times H}\sum_{x=1}^{W/2}\sum_{y=1}^{H}
\mathopen|I_{x,y}^{pred} - I_{W-(x-1),y}^{pred}\mathclose|
\end{split}
\label{eq:sym}
\end{equation}

For simplicity, 
we selectively flip the input so that the occluded part are all on the right side. Besides, only the occluded part (right side) of $I^{pred}$ receives the symmetry loss, i.e. we explicitly pull the right side to be closer to the left.
$L_{sym}$'s contribution is twofold, generating realistic images by encouraging a symmetrical structure and accelerating the convergence of TP-GAN by providing additional back-propagation gradient to relieve self-occlusion for extreme poses.
However, 
due to illumination changes or intrinsic texture difference, pixel values are not strictly symmetric most of the time. Fortunately, 
the pixel difference inside a local area is consistent, and the gradients of a point along all directions are largely reserved under different illuminations. Therefore, the Laplacian space is more robust to illumination changes and more indicative for face structure.


\subsubsection{Adversarial Loss}
The loss for distinguishing real frontal face images $I^{F}$ from synthesized frontal face images $G_{\theta_G}(I^{P})$ is calculated as follows:
\begin{equation}
L_{adv} = \frac{1}{N} \sum_{n=1}^{N} - \log D_{\theta_D}(G_{\theta_G}(I^{P}_n))
\end{equation}
$L_{adv}$ serves as a supervision to push the synthesized image to reside in the manifold of frontal view images. It can prevent blur effect and produce visually pleasing results.

\subsubsection{Identity Preserving Loss}
Preserving the identity while synthesizing the frontal view image is the most critical part in developing the ``recognition via generation'' framework. In this work, we exploit the perceptual loss~\cite{johnson2016perceptual} that is originally proposed for maintaining perceptual similarity to help our model gain the identity preserving ability. Specifically, we define the identity preserving loss based on the activations of the last two layers of the Light CNN~\cite{wulight}:


\begin{equation}
L_{ip} = \sum_{i=1}^{2}\frac{1}{W_i\times H_i} \sum_{x=1}^{W_i}\sum_{y=1}^{H_i}
\mathopen|F(I^{P})^i_{x,y} - F(G(I^{pred}))^i_{x, y}\mathclose|\\
\end{equation}
where $W_i$, $H_i$ denotes the spatial dimension of the last $i$th layer. The identity preserving loss enforces the prediction to have a small distance with the ground-truth in the compact deep feature space. Since the Light CNN is pre-trained to classify tens of thousands of identities, it can capture the most prominent feature or face structure for identity discrimination. Therefore, it is totally viable to leverage this loss to enforce an identity preserving frontal view synthesis.

$L_{ip}$ has better performance when used with $L_{adv}$. Using $L_{ip}$ alone makes the results prone to annoying artifacts, because the search for a local minimum of $L_{ip}$ may go through a path that resides outside the manifold of natural face images. Using $L_{adv}$ and $L_{ip}$ together can ensure that the search resides in that manifold and produces photorealistic image. 

\subsubsection{Overall Objective Function}
The final synthesis loss function is a weighted sum of all the losses defined above:
\begin{equation}
\begin{split}
L_{syn} = L_{pixel} + \lambda_1L_{sym} + \lambda_2L_{adv} + \lambda_3L_{ip} + \lambda_4L_{tv}
\label{eq:trainingLoss}
\end{split}
\end{equation}
We also impose a total variation regularization $L_{tv}$ \cite{johnson2016perceptual} on the synthesized result to reduce spike artifacts.

\section{Experiments}

\begin{figure*}[t]
\centering
\includegraphics[width=0.8\linewidth]{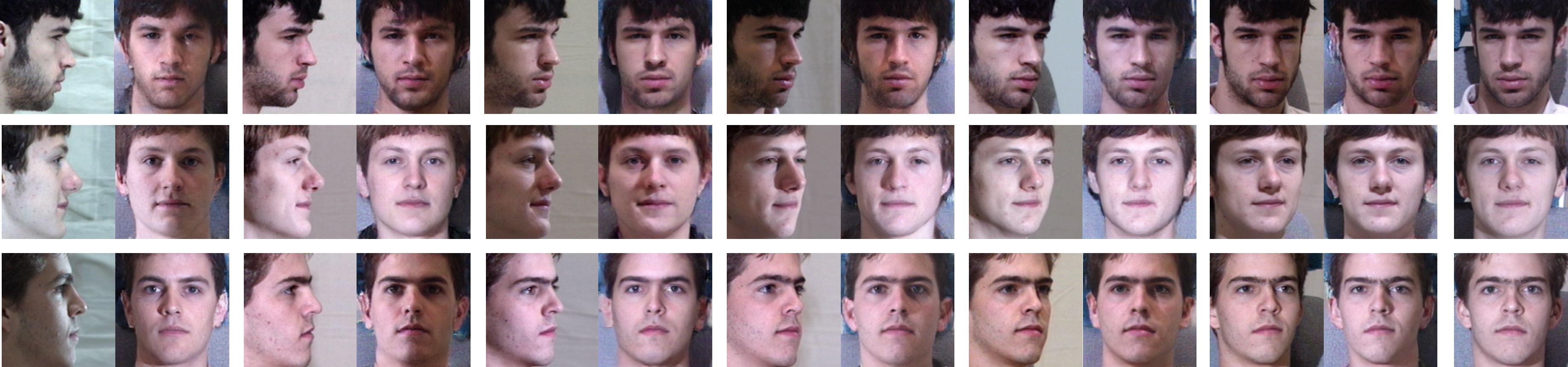}
   \caption{Synthesis results by TP-GAN under different poses. From left to right, the poses are $90^\circ$,  $75^\circ$, $60^\circ$, $45^\circ$, $30^\circ$ and $15^\circ$. The ground truth frontal images are provided at the last column.}
\label{fig:ownsynthesis}
\end{figure*}

\begin{figure}[t]
\centering
\includegraphics[width=0.75\linewidth]{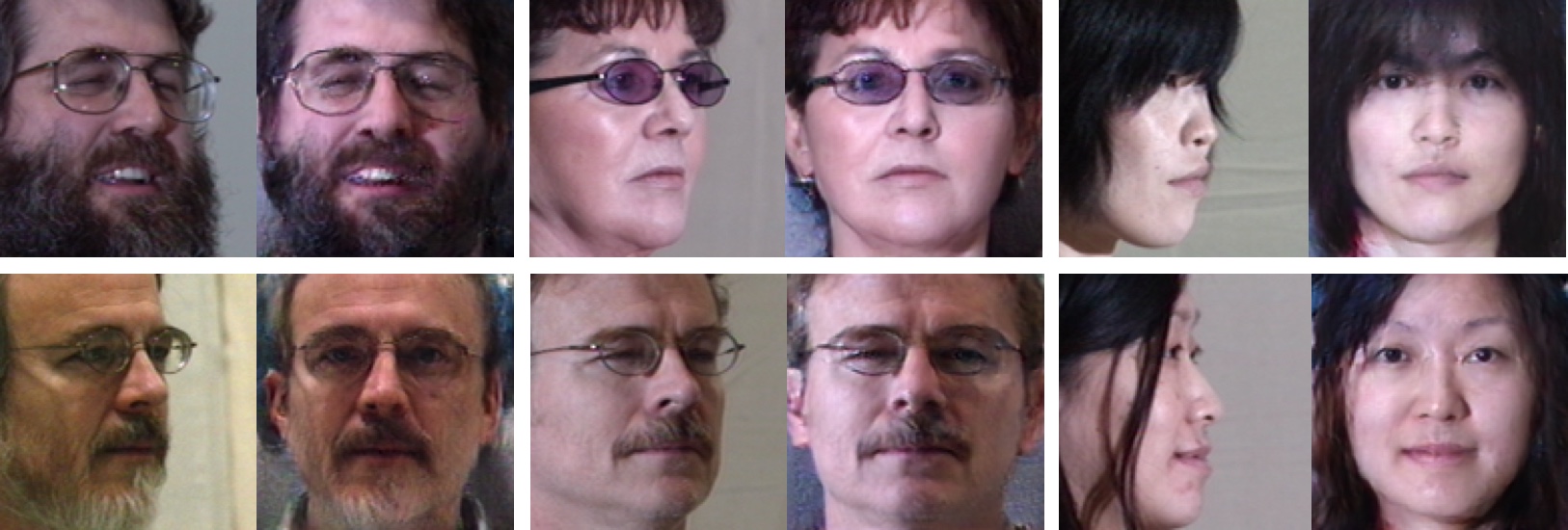}
   \caption{Challenging situations. The facial attributes, e.g. beard, eyeglasses are preserved by TP-GAN. The occluded forehead and cheek are recovered.}
\label{fig:difficulty}
\end{figure}

Except for synthesizing natural looking frontal view images, the proposed TP-GAN also aims to generate identity preserving image for accurate face analysis with off-the-shelf deep features. Therefore, in this section, we demonstrate the merits of our model on qualitative synthesis results and quantitive recognition results in Sec.~\ref{subsec:facesynthesis} and~\ref{subsec:identity}. Sec.~\ref{subsec:visualization} presents visualization of the final deep feature representations to illustrate the effectiveness of TP-GAN. Finally, in Sec.~\ref{subsec:comparison}, we conduct detailed algorithmic evaluation to demonstrate the advantages of the proposed two-pathway architecture and synthesis loss function.

\noindent \textbf{Implementation details} We use colorful images of size $128 \times 128 \times 3$ in all our experiments for both the input $I^{P}$ and the prediction $I^{pred} = G_{\theta_G}(I^{P})$. Our method is evaluated on MultiPIE\cite{multipie}, a large dataset with $750,000+$ images for face recognition under pose, illumination and expression changes. The feature extraction network, Light CNN, is trained on MS-Celeb-1M\cite{msceleb} and fine-tuned on the original images of MultiPIE. Our network is implemented with Tensorflow \cite{tf}. The training of TP-GAN lasts for one day with a batch size of $10$ and a learning rate of $10^{-4}$. In all our experiments, we empirically set $\alpha=10^{-3}, \lambda_1=0.3, \lambda_2 = 10^{-3}, \lambda_3 = 3\times10^{-3}$ and $\lambda_4 = 10^{-4}$.
 \begin{table}\scriptsize
\centering
\caption{Rank-1 recognition rates ($\%$) across views and illuminations under Setting 1. For all the remaining tables, only methods marked with * follow the ``recognition via generation" procedure while others leverage intermediate features for face recognition.}

\begin{tabular}{lclclclclcl} \hline
 Method&$\pm90^\circ$&$\pm75^\circ$&$\pm60^\circ$&$\pm45^\circ$&$\pm30^\circ$&$\pm15^\circ$\\
 \hline
 \hline
 CPF \cite{yim2015rotating} & - & - & - & 71.65 & 81.05 & 89.45\\
 Hassner \etal*\cite{hassner2015effective}  & - & - & 44.81 & 74.68 & 89.59 & 96.78\\
 HPN \cite{Ding2016} & 29.82 & 47.57 & 61.24 & 72.77 & 78.26 & 84.23\\
 FIP\_40\cite{zhu2013deep} & 31.37 & 49.10 & 69.75 & 85.54 & 92.98 & 96.30\\
 c-CNN Forest\cite{modality} & 47.26 & 60.66 & 74.38 & 89.02 & 94.05 & 96.97\\
  \hline
 Light CNN\cite{wulight} & 9.00 & 32.35 & 73.30 & 97.45 & 99.80 & 99.78\\
 TP-GAN* & \bf{64.03} & \bf{84.10} & \bf{92.93} & \bf{98.58} & \bf{99.85} & \bf{99.78}\\
   \end{tabular}\label{tab:set1}
\end{table}




\subsection{Face Synthesis}\label{subsec:facesynthesis}

Most of the previous work on frontal view synthesis are dedicated to address that problem within a pose range of $\pm60^\circ$. Because it is commonly believed that with a pose larger than $60^\circ$, it is difficult to faithfully recover a frontal view image. However, we will show that given enough training data and a proper architecture and loss design, it is in fact feasible to recover photorealistic frontal views from very large poses. Fig.~\ref{fig:ownsynthesis} shows TP-GAN's ability to recover compelling identity-preserving frontal faces from any pose and Fig.~\ref{fig:othermethods} illustrates a comparison with state-of-the-art face frontalization methods. Note that most of TP-GAN's competitors cannot deal with poses larger than $45^\circ$, therefore, we only report their results under $30^\circ$ and $45^\circ$.

\begin{figure}[t]
\centering
\subfigure[Ours]{
\begin{minipage}[b]{0.12\linewidth}
    \centering
   \includegraphics[width=1\linewidth]{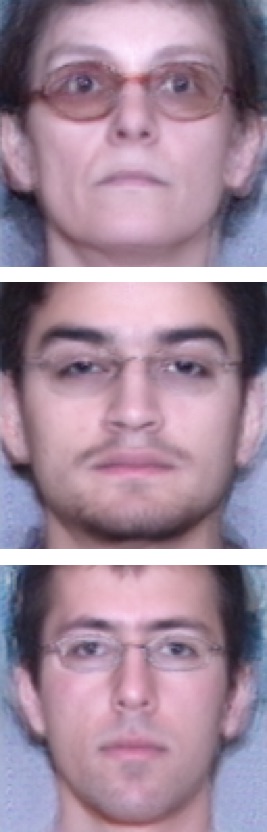}\vspace{0.03\linewidth}
\end{minipage}}
\subfigure[~\cite{yim2015rotating}]{
\begin{minipage}[b]{0.12\linewidth}
    \centering
   \includegraphics[width=1\linewidth]{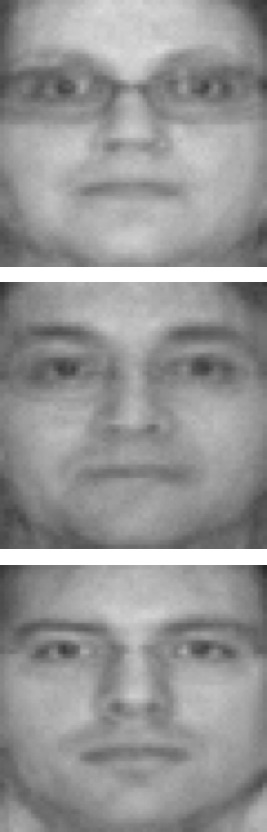}\vspace{0.03\linewidth}
\end{minipage}}
\subfigure[~\cite{face_bmvc16}]{
\begin{minipage}[b]{0.12\linewidth}
    \centering
   \includegraphics[width=1\linewidth]{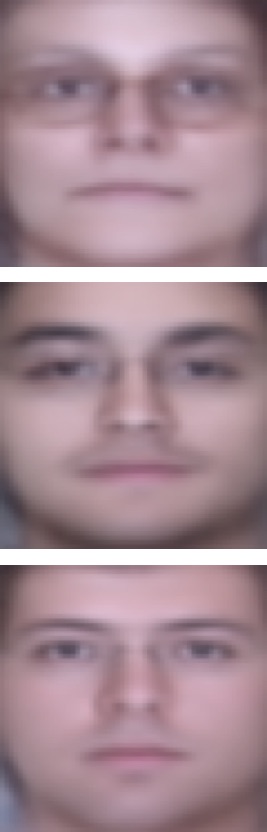}\vspace{0.03\linewidth}
\end{minipage}}
\subfigure[~\cite{zhu2015high}]{
\begin{minipage}[b]{0.12\linewidth}
    \centering
   \includegraphics[width=1\linewidth]{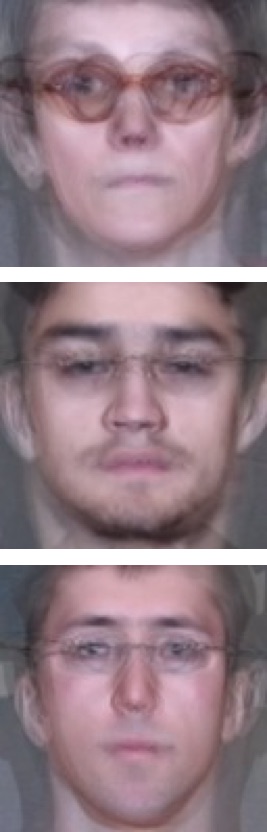}\vspace{0.03\linewidth}
\end{minipage}}
\subfigure[~\cite{hassner2015effective}]{
\begin{minipage}[b]{0.12\linewidth}
    \centering
   \includegraphics[width=1\linewidth]{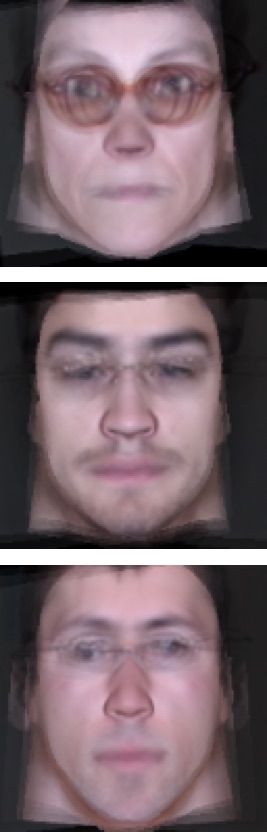}\vspace{0.03\linewidth}
\end{minipage}}
\caption{Mean faces from six images (within $\pm45^{\circ}$) per identity.}
\label{fig:mean} 
\end{figure}


Compared to competing methods, TP-GAN presents a good identity preserving quality while producing photorealistic synthesis. Thanks to the data-driven modeling with prior knowledge from $L_{adv}$ and $L_{ip}$, not only the overall face structure but also the occluded ears, cheeks and forehead can be hallucinated in an identity consistent way. Moreover, it also perfectly preserves observed face attributes in the original profile image, \eg eyeglasses and hair style, as shown in Fig. \ref{fig:difficulty}. 






To further demonstrate the stable geometry shape of the syntheses across multiple poses, we show the mean image of synthesized faces from different poses in Fig. \ref{fig:mean}. The mean faces from TP-GAN preserve more texture detail and contain less blur effect, showing a stable geometry shape across multiple syntheses. Note that our method does not rely on any 3D knowledge for geometry shape estimation, the inference is made through sheer data-driven learning.

\begin{figure}[t]
\centering
\subfigure[LFW]{
\begin{minipage}[b]{0.12\linewidth}
    \centering
   \includegraphics[width=1\linewidth]{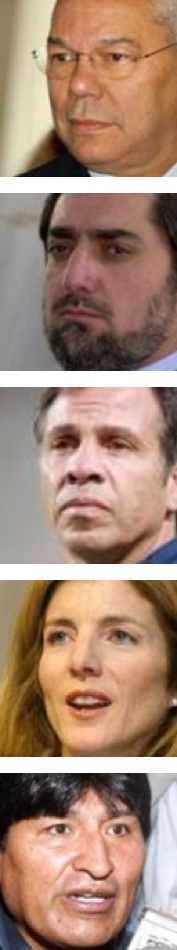}\vspace{0.03\linewidth}
\end{minipage}}
\subfigure[Ours]{
\begin{minipage}[b]{0.12\linewidth}
    \centering
   \includegraphics[width=1\linewidth]{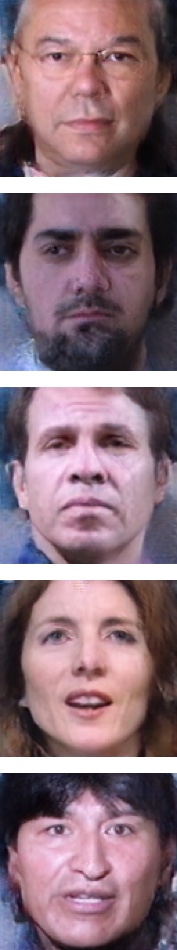}\vspace{0.03\linewidth}
\end{minipage}}
\subfigure[~\cite{zhu2015high}]{
\begin{minipage}[b]{0.12\linewidth}
    \centering
   \includegraphics[width=1\linewidth]{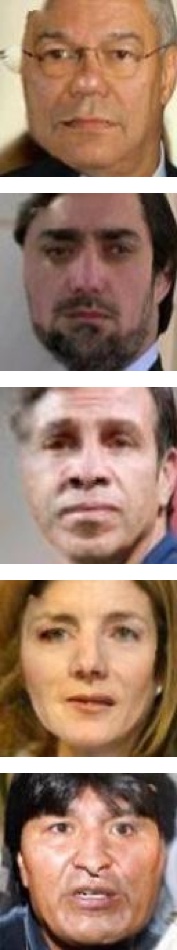}\vspace{0.03\linewidth}
\end{minipage}}
\subfigure[~\cite{hassner2015effective}]{
\begin{minipage}[b]{0.12\linewidth}
    \centering
   \includegraphics[width=1\linewidth]{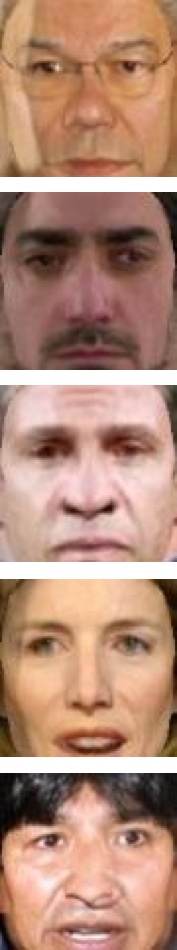}\vspace{0.03\linewidth}
\end{minipage}}
\caption{Synthesis results on the LFW dataset. Note that TP-GAN is trained on Mulit-PIE.}
\label{fig:lfw} 
\end{figure}

 As a demonstration of our model's superior generalization ability to in the wild faces, we use images from LFW~\cite{LFWTech} dataset to test a TP-GAN model trained solely on Multi-PIE. As shown in Fig.~\ref{fig:lfw}, although the resultant color tone is similar to images from Multi-PIE, TP-GAN can faithfully synthesize frontal view images with both finer details and better global shapes for faces in LFW dataset compared to state-of-the-art methods like~\cite{zhu2015high, hassner2015effective}.



\subsection{Identity Preserving Property}\label{subsec:identity}
\noindent \textbf{Face Recognition} To quantitatively demonstrate our method's identity preserving ability, we conduct face recognition on MultiPIE with two different settings. The experiments are conducted by firstly extracting deep features with Light-CNN \cite{wulight} and then compare Rank-1 recognition accuracy with a cosine-distance metric. The results on the profile images $I^{P}$ serve as our baseline and are marked by the notation Light-CNN in all tables. It should be noted that although many deep learning methods have been proposed for frontal view synthesis, none of their synthesized images proved to be effective for recognition tasks. In a recent study on face hallucination~\cite{wu2016deep}, the authors show that directly using a CNN synthesized high resolution face image for recognition will certainly degenerate the performance instead of improving it. Therefore, it is of great significance to validate whether our synthesis results can boost the recognition performance (whether the ``recognition via generation" procedure works). 

In Setting 1, we follow the protocol from \cite{modality}, and only images from session one are used. We include images with neutral expression under 20 illuminations and 11 poses within $\pm 90^\circ$. One gallery image with frontal view and illumination is used for each testing subject. There is no overlap between training and testing sets. Table~\ref{tab:set1} shows our recognition performance and the comparison with the state-of-the-art. TP-GAN consistently achieves the best performance across all angles, and the larger the angle, the greater the improvement. When compared with c-CNN Forest\cite{modality}, which is an ensemble of three models, we achieve a performance boost of about 20\% on large pose cases. 


\begin{table}\scriptsize
\centering
\caption{Rank-1 recognition rates ($\%$) across views, illuminations and sessions under Setting 2.}
\begin{tabular}{lclclclclcl} \hline
 Method&$\pm90^\circ$&$\pm75^\circ$&$\pm60^\circ$&$\pm45^\circ$&$\pm30^\circ$&$\pm15^\circ$\\
 \hline \hline
 FIP+LDA\cite{zhu2013deep} & - & - & 45.9 & 64.1 & 80.7 & 90.7\\
MVP+LDA\cite{zhu2014multi} & - & -  & 60.1 & 72.9 & 83.7 & 92.8 \\
CPF \cite{yim2015rotating} & - & - & 61.9 & 79.9 & 88.5 & 95.0\\
 DR-GAN \cite{tran17}  & - & - & 83.2 & 86.2 & 90.1 & 94.0\\
  \hline
  Light CNN\cite{wulight} & 5.51 & 24.18 & 62.09 & 92.13 & 97.38 & 98.59\\
 TP-GAN* & \bf{64.64} & \bf{77.43} & \bf{87.72} & \bf{95.38} & \bf{98.06} & \bf{98.68}\\
   \end{tabular}\label{tab:set2}
\end{table}

\begin{table}\scriptsize
\centering
\caption{Gender classification accuracy ($\%$) across views and illuminations.}
\begin{tabular}{lclclclclclcl} \hline
 Method&$\pm45^\circ$&$\pm30^\circ$&$\pm15^\circ$\\
 \hline
 \hline
 $I^{P}_{60}$ & 85.46 & 87.14 & 90.05\\
 CPI* \cite{yim2015rotating} & 76.80 & 78.75 & 81.55\\
Amir \etal * \cite{face_bmvc16} & 77.65 & 79.70 & 82.05\\
\hline
$I^{P}_{128}$ & 86.22 & 87.70 & 90.46\\
Hassner \etal*~\cite{hassner2015effective} & 83.83 & 84.74 & 87.15 \\
TP-GAN* & \bf{90.71} & \bf{89.90} & \bf{91.22}\\
   \end{tabular}\label{tab:gender}
\end{table}

In Setting 2, we follow the protocol from \cite{yim2015rotating}, where neural expression images from all four sessions are used. One gallery image is selected for each testing identity from their first appearance. All synthesized images of MultiPIE in this paper are from the testing identities under Setting 2.  The result is shown in Table~\ref{tab:set2}. 
Note that all the compared CNN based methods achieve their best performances with learned intermediate features, whereas we directly use the synthesized images following a ``recognition via generation" procedure. 

\noindent \textbf{Gender Classification} To further demonstrate the potential of our synthesized images on other facial analysis tasks, we conduct an experiment on gender classification. All the compared methods in this part also follow the ``recognition via generation" procedure, where we directly use their synthesis results for gender classification. The CNN for gender classification is of the same structure as the encoder $G_{\theta ^g_E}$ and is trained on $batch1$ of the UMD\cite{umdfaces} dataset. 

We report the testing performance on Multi-PIE (Setting-1) in Table \ref{tab:gender}.  For fair comparison, we present the results on the unrotated original images in two resolutions, $128\times 128$ ($I^{P}_{128}$) and $60\times60$ ($I^{P}_{60}$) respectively. TP-GAN's synthesis achieves a better classification accuracy than the original profile images due to normalized views. It's not surprising to see that all other compared models perform worse than the baseline, as their architectures are not designed for the gender classification task. Similar phenomenon is observed in~\cite{wu2016deep} where synthesized high resolution face images severely degenerate the recognition performance instead of improving it. That indicates the high risk of losing prominent facial features of $I^{P}$ when manipulating images in the pixel space. 

\subsection{Feature Visualization}\label{subsec:visualization}
%

\begin{figure}[t]
\centering
\includegraphics[width=0.8\linewidth]{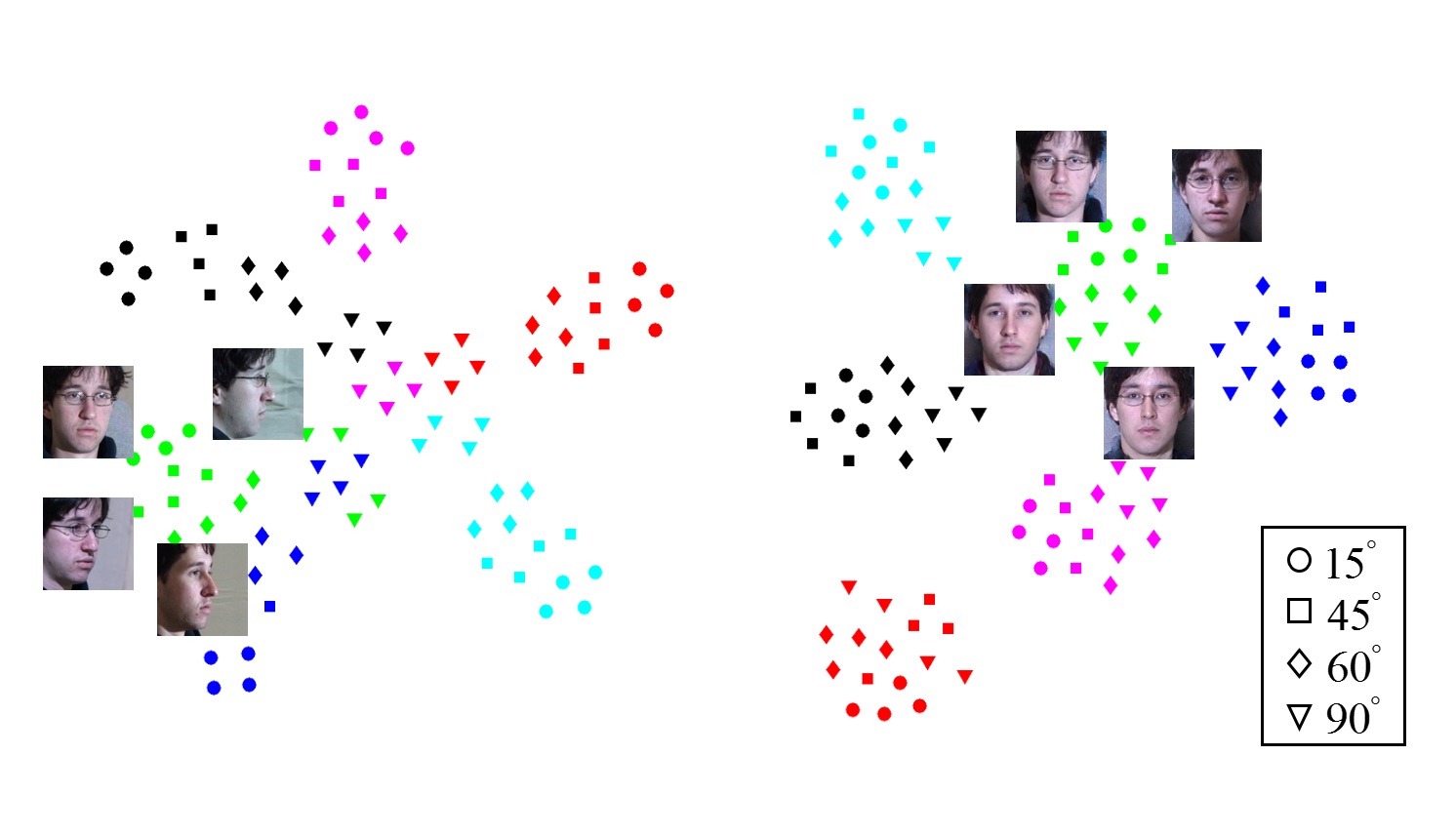}
   \caption{Feature space of the profile faces (left) and fontal view synthesized images (right). Each color represents a different identity. Each shape represent a view. The images for one identity are labeled.}
\label{fig:tsne}
\end{figure}

We use t-SNE\cite{tsne} to visualize the 256-dim deep feature on a two dimensional space. The left side of Fig.~\ref{fig:tsne} illustrates the deep feature space of the original profile images. It's clear that images with a large pose ($90^\circ$ in particular) are not separable in the deep feature space spanned by the Light-CNN. It reveals that even though the Light-CNN is trained with millions of images, it still cannot properly deal with large pose face recognition problems. On the right side, after frontal view synthesis with our TP-GAN, the generated frontal view images can be easily classified into different groups according to their identities.
\begin{table}\scriptsize
\centering
\caption{Model comparison: Rank-1 recognition rates ($\%$) under Setting 2.}
\begin{tabular}{lclclclclclcl} \hline
 Method&$\pm90^\circ$&$\pm75^\circ$&$\pm60^\circ$&$\pm45^\circ$&$\pm30^\circ$&$\pm15^\circ$\\
 \hline \hline
w/o P &44.13& 66.10 & 80.64 & 92.07 & 96.59 & 98.35 \\
w/o $L_{ip}$ &43.23 & 56.55 & 70.99 & 85.87 & 93.43 & 97.06\\
w/o $L_{adv}$ &62.83 & 76.10 & 85.04 & 92.45 & 96.34 & 98.09\\
w/o $L_{sym}$ &62.47 & 75.71 & 85.23 & 93.13 & 96.50 & 98.47\\
TP-GAN& \bf{64.64} & \bf{77.43} & \bf{87.72} & \bf{95.38} & \bf{98.06} & \bf{98.68}\\
   \end{tabular}\label{tab:compare}
\end{table}

\subsection{Algorithmic analysis}\label{subsec:comparison}
In this section, we go over different architectures and loss function combinations to gain insight into their respective roles in frontal view synthesis. Both qualitative visualization results and quantitive recognition results are reported for a comprehensive comparison.

We compare four variations of TP-GAN in this section, one for comparing the architectures and the other three for comparing the objective functions. Specifically, we train a network without the local pathway (denoted as P) as the first variant. 
With regards to the loss function, we keep the two-pathway architecture intact and remove one of the three losses, \ie $L_{ip}, L_{adv}$ and $L_{sym}$, in each case.

\begin{figure}[t]
\centering
\subfigure[methods]{
\begin{minipage}[b]{0.16\linewidth}
   \includegraphics[width=1\linewidth]{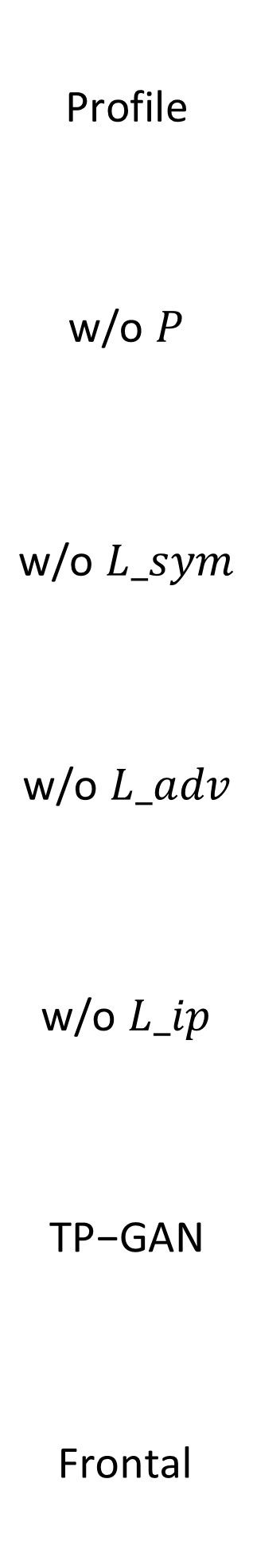}\vspace{0.03\linewidth}
\end{minipage}}
\subfigure[$90^\circ$]{
\begin{minipage}[b]{0.138\linewidth}
    \centering
   \includegraphics[width=1\linewidth]{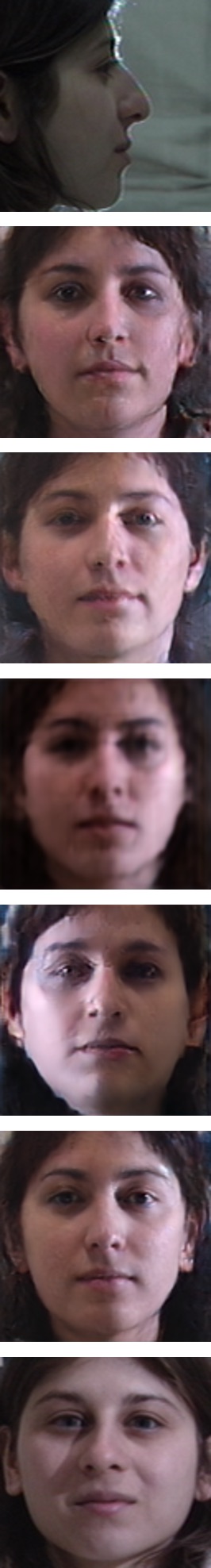}\vspace{0.03\linewidth}
\end{minipage}}
\subfigure[$75^\circ$]{
\begin{minipage}[b]{0.14\linewidth}
    \centering
   \includegraphics[width=1\linewidth]{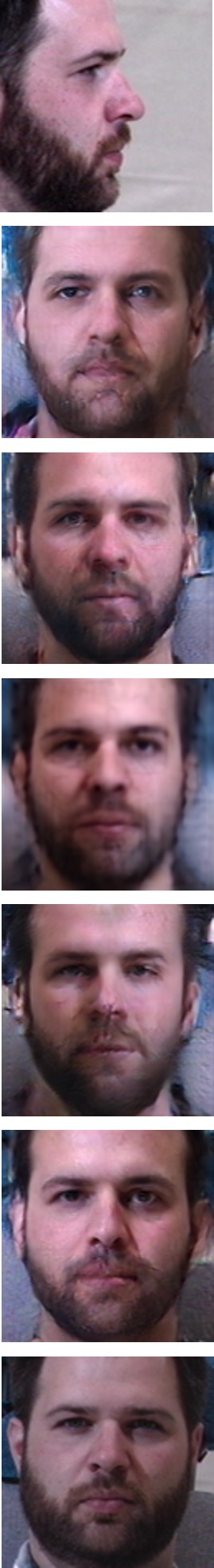}\vspace{0.03\linewidth}
\end{minipage}}
\subfigure[$60^\circ$]{
\begin{minipage}[b]{0.139\linewidth}
    \centering
   \includegraphics[width=1\linewidth]{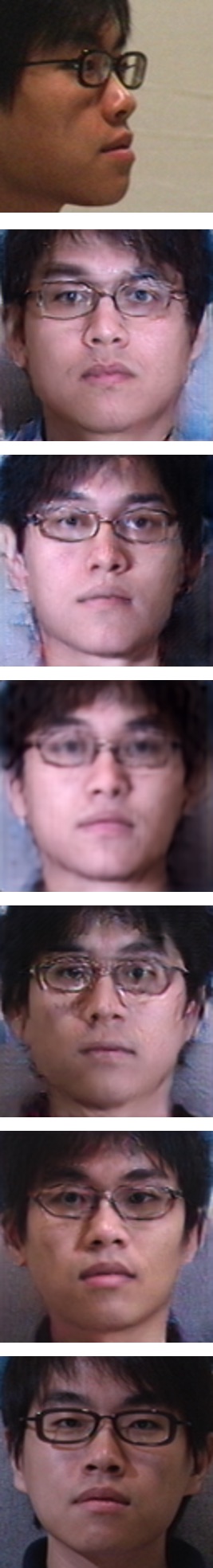}\vspace{0.03\linewidth}
\end{minipage}}
\subfigure[$30^\circ$]{
\begin{minipage}[b]{0.139\linewidth}
    \centering
   \includegraphics[width=1\linewidth]{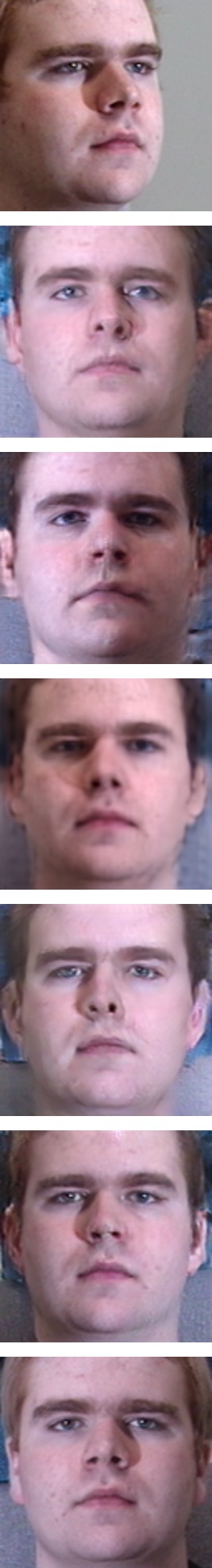}\vspace{0.03\linewidth}
\end{minipage}}

\caption{Model comparison: synthesis results of TP-GAN and its variants.}
\label{fig:modelcomp} 
\end{figure}

Detailed recognition performance is reported in Table~\ref{tab:compare}. The two-pathway architecture and the identity preserving loss contribute the most for improving the recognition performance, especially on large pose cases. Although not as much apparent, both the symmetry loss and the adversarial loss help to improve the recognition performance. Fig.~\ref{fig:modelcomp} illustrates the perceptual performance of these variants. As expected, inference results without the identity preserving loss or the local pathway deviate from the true appearance seriously. And the synthesis without adversarial loss tends to be very blurry, while the result without the symmetry loss sometimes shows unnatural asymmetry effect. 


\section{Conclusion}
In this paper, we have presented a global and local perception GAN framework for frontal view synthesis from a single image. The framework contains two separate pathways, modeling the out-of-plane rotation of the global structure and the non-linear transformation of the local texture respectively. 
To make the ill-posed synthesis problem well constrained, we further introduce adversarial loss, symmetry loss and identity preserving loss in the training process. Adversarial loss can faithfully discover and guide the synthesis to reside in the data distribution of frontal faces. Symmetry loss can explicitly exploit the symmetry prior to ease the effect of self-occlusion in large pose cases. Moreover, identity preserving loss is incorporated into our framework, so that the synthesis results are not only visually appealing but also readily applicable to accurate face recognition. Experimental results demonstrate that our method not only presents compelling perceptual results but also outperforms state-of-the-art results on large pose face recognition.

\section{Supplementary Material}
\subsection{Detailed Network Architecture}\label{sec:arch}
\begin{figure*}[th]
\centering
\includegraphics[width=0.9\linewidth]{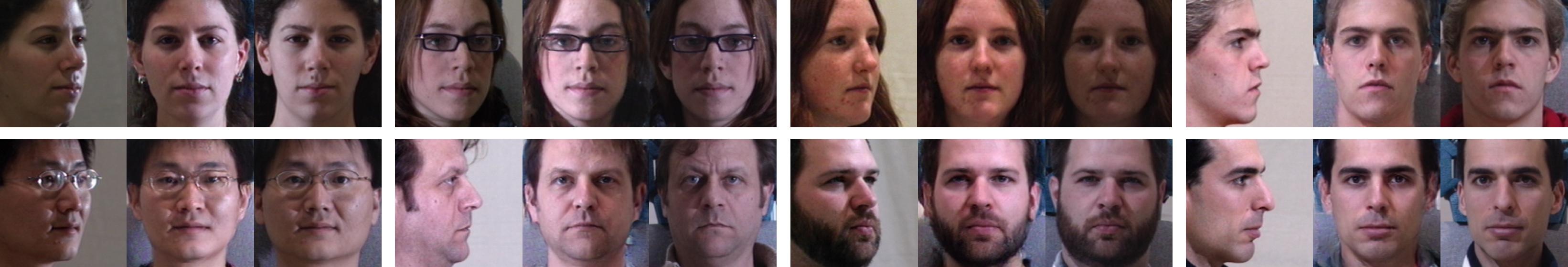}
\caption{Our synthesized images present moderately better exposure in some cases. Each tuple consists of three images, with the input $I^P$ on the left, the synthesized in the middle, the ground truth frontal face $I^{gt}$ on the right. Each $I^P$ and its corresponding $I^{gt}$ are taken under a flash light from the same direction.}
\label{fig:illum2}
\end{figure*}
\begin{figure*}[t]
\centering
\includegraphics[width=0.7\linewidth]{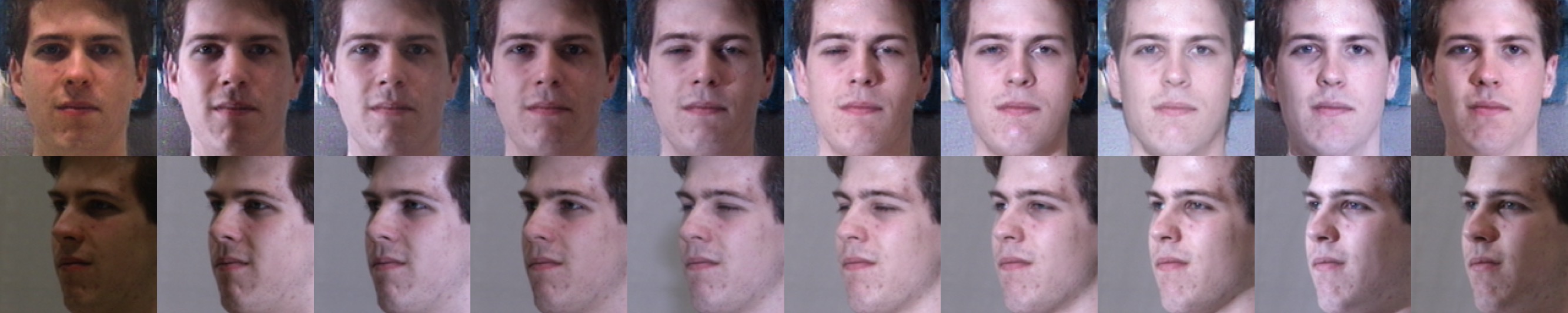}
   \caption{Synthesis results under various illuminations. The first row is the synthesized image, the second row is the input. Please to refer to the supplementary material for more results. }
\label{fig:illumination}
\end{figure*}

The detailed structures of the global pathway $G_{\theta ^g_E}$ and $G_{\theta ^g_D}$ are provided in Table~\ref{tab:globalE} and Table~\ref{tab:globalD}. Each convolution layer of $G_{\theta ^g_E}$ is followed by one residual block\cite{residule}. Particularly, the layer $conv4$ is followed by four blocks. The output of the layer $fc2$ ($v_{id}$) is obtained by selecting the maximum element from the two split halves of $fc1$.

The Decoder of the global pathway  $G_{\theta ^g_D}$ contains two parts. The first part is a simple deconvolution stack for up-sampling the concatenation of the feature vector $v_{id}$ and the random noise vector $z$. 
The second part is the main deconvolution stack for reconstruction. Each layer takes the output of its previous layer as the regular input, which is omitted in the table for readability. Any extra inputs are specified in the $Input$ column. Particularly, the layers $feat8$ and $deconv0$ have their complete inputs specified. Those extra inputs instantiate the skipping layers and the bridge between the two pathways. The fused feature tensor from the local pathway is denoted as $local$ in Table~\ref{tab:globalD}. Tensor $local$ is the fusion of the outputs of four $G_{\theta ^l_D}$s' layer $conv4$ (of Table~\ref{tab:local}). To mix the information of the various inputs, all extra inputs pass through one or two residual blocks before being concatenated for deconvolution.   The profile image $I^{P}$ is resized to the corresponding resolution and provides a shortcut access to the original texture for $G_{\theta ^g_D}$.

Table~\ref{tab:local} shows the structures of the local pathway $G_{\theta ^l_E}$ and  $G_{\theta ^l_D}$. The local pathway contains three down-sampling and up-sampling processes respectively. The $w$ and $h$ denote the width and the height of the cropped patch. For the patches of the two eyes, we set $w$ and $h$ as 40; for the patch of the nose, we set $w$ as 40 and $h$ as 32; for the patch of the mouth, we set $w$ and $h$ as 48 and 32 respectively.

We use rectified linear units (ReLU) \cite{relu} as the non-linearity activation and adopt batch normalization \cite{batch} except for the last layer. In $G_{\theta ^g_E}$ and $G_{\theta ^l_E}$, the leaky ReLU is adopted.
\begin{table}\scriptsize
\centering
\caption{Structure of the Encoder of the global pathway $G_{\theta ^g_E}$}
\begin{tabular}{|c|c|c|} \hline
Layer&Filter Size&Output Size\\
\hline \hline
conv0 & $7\times7 / 1$ & $128 \times 128 \times 64$\\
conv1 & $5\times5 / 2$ & $64 \times 64 \times 64$\\
conv2 & $3\times3 / 2$ & $32 \times 32 \times 128$\\
conv3 & $3\times3 / 2$ & $16 \times 16 \times 256$\\
conv4 & $3\times3 / 2$ & $8 \times 8 \times 512$\\
\hline
fc1 & - & 512\\
fc2 & - & 256\\
\hline
\end{tabular}\label{tab:globalE}
\end{table}

\begin{table}\scriptsize
\centering
\caption{Structure of the Decoder of the global pathway $G_{\theta ^g_D}$.
The $conv$s in $Input$ column refer to those in Table~\ref{tab:globalE}.}
\begin{tabular}{|c|c|c|c|} \hline
Layer&Input&Filter Size&Output Size\\
\hline \hline
feat8& fc2, z & - & $8 \times 8 \times 64$ \\
feat32& - & $3\times3 / 4$ & $32 \times 32 \times 32$ \\
feat64& - & $3\times3 / 2$ & $64 \times 64 \times 16$ \\
feat32& - & $3\times3 / 2$ & $128 \times128 \times 8$ \\
\hline
deconv0 & feat8, conv4 & $3\times3 / 2$ & $16 \times 16 \times 512$\\
deconv1 & conv3 & $3\times3 / 2$ & $32 \times 32 \times 256$\\
deconv2 &feat32, conv2, $I^P$& $3\times3 / 2$ & $64 \times 64 \times 128$\\
deconv3 &feat64, conv1, $I^P$& $3\times3 / 2$ & $128 \times 128 \times 64$\\
conv5 & feat128, conv0,  $local$ , $I^P$ & $5\times5 / 1$ & $128 \times128 \times 64$\\
conv6 & - &  $3\times3 / 1$ & $128 \times128 \times 32$\\
conv7 & - &  $3\times3 / 1$ & $128 \times128 \times 3$\\
\hline
\end{tabular}\label{tab:globalD}
\end{table}

\begin{table}\scriptsize
\centering
\caption{Structure of the local pathway $G_{\theta ^l_E}$ \& $G_{\theta ^l_D}$.
The $conv$s in $Input$ column refer to those in the same table.}
\begin{tabular}{|c|c|c|c|} \hline
Layer&Input&Filter Size&Output Size\\
\hline \hline
conv0& - &  $3\times3 / 1$ & $w \times h \times 64$ \\
conv1& - & $3\times3 / 2$ & $w/2 \times h/2  \times 128$ \\
conv2& - & $3\times3 / 2$ & $w/4 \times h/4  \times 256$ \\
conv3& - & $3\times3 / 2$ & $w/8 \times h/8  \times 512$ \\
\hline
deconv0 & conv3 & $3\times3 / 2$ & $w/4 \times h/4 \times 256$\\
deconv1 & conv2 & $3\times3 / 2$ & $w/2 \times h/2 \times 128$\\
deconv2 & conv1 &  $3\times3 / 2$ & $ w \times h \times 64$\\
conv4 & conv0 & $3\times3 / 1$ &  $ w \times h  \times 64$\\
conv5 & - &  $3\times3 / 1$ &  $ w \times h \times 3$\\

\hline
\end{tabular}\label{tab:local}
\end{table}

\textbf{Discussion:} 
Our model is simple while achieving better performance in terms of the photorealism of synthesized images.
 Yim \etal \cite{yim2015rotating} and Zhu \etal \cite{zhu2013deep} use locally connected convolutional layers for feature extraction and fully connected layer for synthesis. We use weight-sharing convolution in most cases. 
Our model reduces parameter numbers to a large extent and avoids expensive computation for generating every pixel during synthesis. 
Yim \etal \cite{yim2015rotating} and Amir \etal \cite{face_bmvc16} add a second reconstruction branch or a refinement network. Our early supervised decoder achieves end-to-end generation of high-resolution image.

\subsection{Additional Synthesis Results}
Additional synthesized images $I^{pred}$ are shown in Fig.~\ref{fig:illum2} and Fig.~\ref{fig:illumination}. 
Under extreme illumination condition, the exposure of $I^{pred}$ is consistent with or moderately better than that of its input $I^{P}$ or its ground truth frontal face $I^{gt}$.
{Fig.~\ref{fig:illumination} demonstrates TP-GAN's robustness to illumination changes. Despite extreme illumination variations, the skin tone, global structure and local details are consistent across illuminations. Our method can automatically adjust $I^{P}$'s exposure and white balance.

Additionally, we use a state-of-the-art face alignment method\cite{hongwen} to provide four landmarks for TP-GAN under extreme poses. The result is only slightly worse than that reported in Table 2 of the paper. Specifically, we achieve Rank-1 recognition rates of 87.63($\pm60^\circ$), 76.69($\pm75^\circ$), 62.43($\pm90^\circ$).

\begin{figure}[t]
\centering
\includegraphics[width=0.8\linewidth]{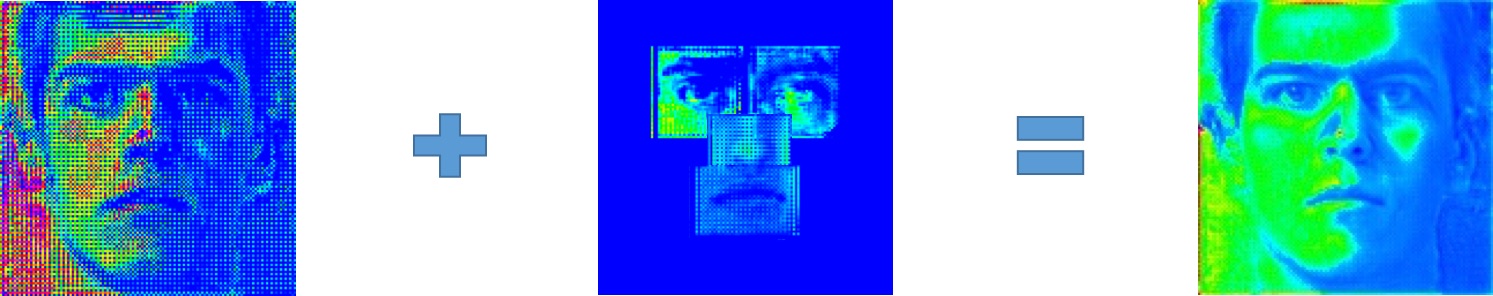}
   \caption{Synthesis process illustrated from the perspective of activation maps. The up-sampled feature map  $C_g$  is combined with the local pathway feature map $C_l$ to produce feature maps with detailed texture.}
\label{fig:activation}
\end{figure}

\begin{figure}[t]
\centering
\includegraphics[width=0.6\linewidth]{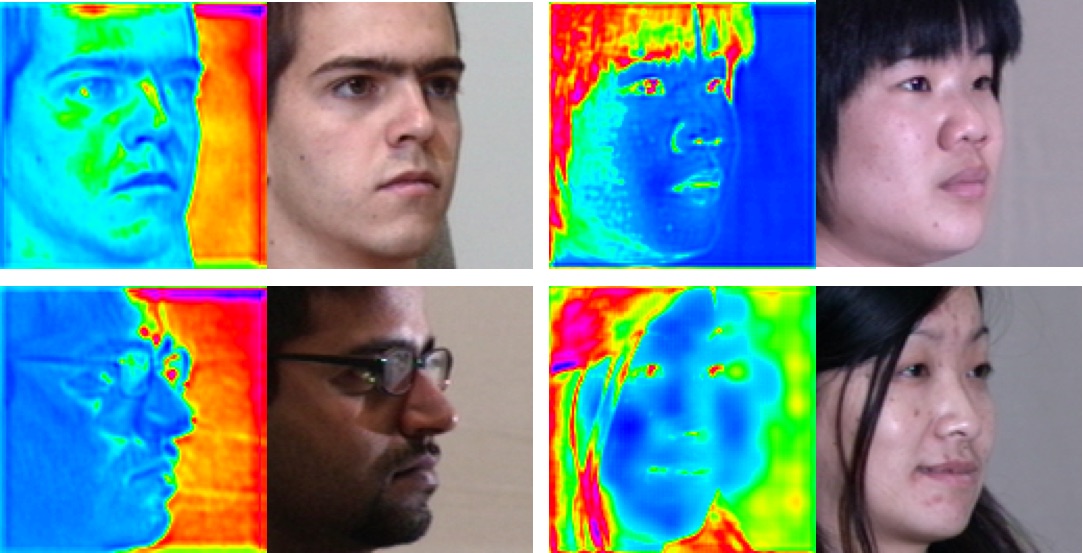}
   \caption{Automatic detection of certain semantic regions. Some skip layers' activation maps are sensitive to certain semantic regions. One for detecting non-face region is shown on the left, another for detecting hair region is shown on the right. Note the delicate and complex region boundaries around the eyeglasses and the fringe.}
\label{fig:seg}
\end{figure}

\subsection{Activation Maps Visualization}\label{subsec:visualization}

In this part, we visualize the intermediate feature maps to gain some insights into the processing mechanism of the two-pathway network. Fig.~\ref{fig:activation} illustrates the fusion of global and local information before the final output. $C_g$ contains the up-sampled outputs of the global pathway and $C_l$ refers to the features maps fused from the four local pathways. Their information is concatenated and further integrated by the following convolutional layers.


We also discovered that TP-GAN can automatically detect certain semantic regions. Fig.~\ref{fig:seg} shows that certain skip layers have high activation for regions such as non-face region and hair region. The detection is learned by the network without supervision. Intuitively, dividing the input image into different semantic regions simplifies the following composition or synthesis of the frontal face.

\section*{Acknowledgement}
This work is partially funded by National Natural Science Foundation of China (Grant No. 61622310, 61473289) and the State Key Development Program (Grant No. 2016YFB1001001). We thank Xiang Wu for useful discussion.
{\small
\bibliographystyle{ieee}
\bibliography{egbib}
}

\end{document}